\documentclass[11pt]{article}

\usepackage[preprint]{acl}

\usepackage{times}
\usepackage{latexsym}

\usepackage[T1]{fontenc}

\usepackage[utf8]{inputenc}
\usepackage{amssymb}
\usepackage{booktabs}
\usepackage{tcolorbox} %
\usepackage{enumitem}

\usepackage{microtype}

\usepackage{inconsolata}

\usepackage{graphicx}

\title{ \textsc{DatedGPT}: Preventing Lookahead Bias in Large Language Models with Time-Aware Pretraining}

\author{Yutong Yan$^{\alpha}$ \quad
Raphael Tang$^{\beta}$ \quad
Zhenyu Gao$^{\alpha}$ \quad
Wenxi Jiang$^{\alpha}$ \quad
Yao Lu$^{\beta}$ \\
$^{\alpha}$Department of Finance, CUHK Business School, The Chinese University of Hong Kong \quad \\
$^{\beta}$Centre for Artificial Intelligence, University College London \\
\texttt{yutong.yan@link.cuhk.edu.hk},
\texttt{r-tang.25@ucl.ac.uk},
\\
\texttt{gaozhenyu@baf.cuhk.edu.hk},
\texttt{wenxijiang@baf.cuhk.edu.hk},
\\
\texttt{yao.lu@cs.ucl.ac.uk}
}

\begin{document}
\maketitle
\begin{abstract}
Large language models pretrained on internet-scale data risk lookahead bias in forecasting tasks, as they may have already seen the true outcome during training.
To address this, we present~\textsc{DatedGPT}, a family of twelve 1.3B-parameter language models trained from scratch on approximately 100 billion tokens each with strict annual data cutoffs spanning 2013 to 2024, together with \textsc{DatedInstruct}, an instruction dataset grounded in each year's documents to prevent leakage during post-training.
The models are competitive with open models of similar scale, and perplexity-based probing confirms that each model's knowledge is bounded by its cutoff year.
On stock return prediction over 61,000 firm-day news headlines, \textsc{DatedGPT}-instruct achieves an annualised Sharpe ratio of $3.20$ under the lookahead-bias-free setup. Lookahead-biased models, whose training data covers the outcome period, add a lookahead premium of $26.4$ b.p.\ per standard deviation, significant at the 1\% level. The series thus enables direct analysis of lookahead bias in financial forecasting.
We provide an interactive web demo that allows users to query and compare responses from models across different cutoff years, available at \url{www.datedgpt.com}.\footnote{A demo video is available \href{https://yutongyan.xyz/files/datedgpt_demo_video.mp4}{here}. 
Model checkpoints are available at \url{hf.co/datedGPT}.}
\end{abstract}

\section{Introduction}

Large language models (LLMs)~\cite{radford2019language,touvron2023llama,yang2025qwen3, xu2026deepseekv4}, pretrained on internet-scale corpora, have demonstrated strong capabilities across a wide range of general-purpose tasks and are increasingly applied to real-world financial forecasting problems \citep{bybee2023ghost, lopez2023can, hansen2024can, jha2024chatgpt}.
However, when an LLM correctly predicts the 2008 stock market crash in a backtest, it is unclear whether the prediction derives from the information available at the time, or the model has simply seen the outcome during pretraining.
This concern is known as \textit{lookahead bias}~\citep{sarkar2024lookahead, gao2025test}, and is closely related to test-set leakage~\citep{xu2025infini, jacovi2023stop}.
This issue is crucial in economic and financial forecasting, where a backtest may otherwise report inflated performance that cannot be realised as expected.

\begin{figure}[t]
    \centering
    \includegraphics[width=0.7\columnwidth]{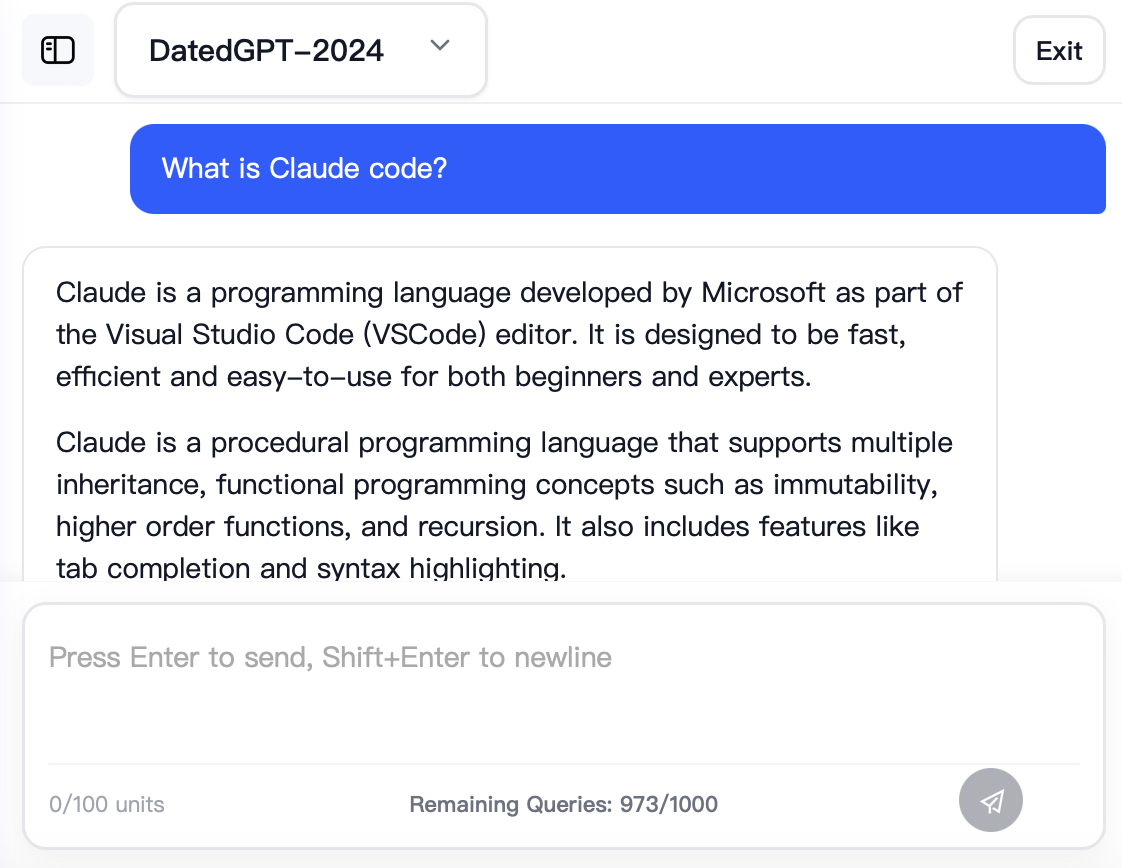}
    \caption{Web interface for \textsc{DatedGPT}. When asked ``What is Claude Code?'' (released February 2025), the \textsc{DatedGPT-2024} model, trained exclusively on data available before 2024, is unaware of the product and hallucinates an incorrect answer.}
    \label{fig:demo}
\end{figure}

The cleanest and most straightforward fix is to train a series of models under a strict knowledge cutoff, so that any prediction dated after the cutoff is free of lookahead bias by construction. 
Existing pretraining efforts, however, move in the opposite direction by expanding data for better performance.

In this work, we propose \textsc{DatedGPT}, a family of twelve 1.3B-parameter language models with explicit annual cutoffs spanning from 2013 to 2024. Each model is trained from scratch on approximately 100B tokens of data available up to its designated cutoff year. To further enhance instruction-following ability, we curate \textsc{DatedInstruct}, a time-aware instruction dataset constructed by mining QA pairs from each year's web data~\citep{yue2024mammoth2}, so that the tuning data likewise contains no information beyond each cutoff date.

Evaluation results on language understanding and instruction-following benchmarks show that our models, despite the modest model size and training token budget, achieve competitive performance with existing models of similar scale. On stock return prediction over 61,000 firm-day news headlines, \textsc{DatedGPT}-instruct achieves an annualised Sharpe ratio of $3.20$ and an economic magnitude of $28.3$~b.p.\ per standard deviation under the lookahead-bias-free setup. Models under the lookahead-biased setup earn a statistically significant premium of $26.4$~b.p.\ per standard deviation ($t = 10.65$) over the lookahead-free setup, showing that lookahead bias is a real concern in financial forecasting. Furthermore, perplexity probing reveals a clear reversal at each model's data cutoff date, confirming that each model's knowledge is indeed bounded by its cutoff year. The \textsc{DatedGPT} series thus offers practical value for financial forecasting tasks. 

It is worth noting that we do not intend to build another financial prediction model to replace frontier proprietary models. Rather, our results show that small-scale models are capable of analysing lookahead bias with statistical significance, and the resulting measure can serve as an auxiliary signal for financial decision systems. This capability comes from the training setup rather than from scale, and cannot be added after the fact to a model that has already seen the future. Our exploration also addresses a long-standing question in the open-model community about what small models from academic labs can contribute beyond a transparent training process.

To summarise, our contributions are as follows:
\begin{itemize}[itemsep=0mm,leftmargin=2ex]
    \item We introduce \textsc{DatedGPT}, a family of 1.3B-parameter language models trained from scratch on temporally partitioned data with annual cutoffs from 2013 to 2024, together with \textsc{DatedInstruct}, a time-aware instruction dataset curated for each cutoff year, both built under a leakage-free setup. To our knowledge, this is among the largest and most complete yearly coverage model series with explicit date cutoffs to date.
    
    \item We evaluate \textsc{DatedGPT} on language understanding benchmarks and pairwise win rates, showing competitive performance at a small scale, while tuning on \textsc{DatedInstruct} consistently improves other time-locked models. On stock return prediction, the models retain predictive power under the lookahead-free setup, and the measured lookahead premium is statistically significant.
    
    \item We release the full model series (twelve models), the \textsc{DatedInstruct} dataset, and a web interface that allows users to compare responses across different cutoff years, supporting temporal evaluation of LLMs.
\end{itemize}

\section{Time-Aware Dataset Curation}

\subsection{Pretraining Data with Cutoff Dates}\label{sec:pretrain-data}
Data quality plays a central role in training performant models, especially at small scale~\citep{touvron2023llama,weber2024redpajama}. We adopt FineWeb-Edu~\citep{penedo2024fineweb} as our base dataset, which \citet{karpathy2024llmc} shows can substantially reduce the compute required to match GPT-2 performance. Each FineWeb-Edu document carries the timestamp at which it was crawled from Common Crawl. Crawl timestamps do not reflect the true creation date of a webpage, as a page crawled in 2015 may have been authored in the 1990s. Filtering by crawl date nevertheless satisfies our primary goal, since it ensures that a model is never exposed to data crawled after its designated cutoff. We therefore filter by crawl date, rounded to the year, and obtain twelve time-aware pretraining sets spanning 2013 to 2024, with approximately 100 billion tokens per year.

\subsection{Instruction Data with Cutoff Dates}\label{sec:instruction-data}

The challenge in curating instruction-tuning data for models with strict date cutoffs is that existing instruction datasets~\citep{weifinetuned, wang2023self,alpaca,zheng2023judging,xuwizardlm} are unsuitable. Their responses are written by human annotators or generated by modern LLMs, and may contain knowledge beyond the cutoff date that is difficult to detect and filter. 
Therefore, we construct the instruction dataset \textsc{DatedInstruct}, built from each year's pretraining documents under a lexical constraint that minimises this risk.

The approach is motivated by extractive text summarisation~\citep{nallapati2017summarunner}, where the output is extracted from the source text and fabricated content is thus excluded, and by instruction mining from web data~\citep{yue2024mammoth2}. Following~\citet{groeneveld2024olmo}, we organise the curated data into four categories: factual knowledge, analytical reasoning, creative writing, and format following. The construction proceeds in two steps.

\paragraph{Source-document selection} Due to the diverse formats and noisy nature of web-crawled data, not every pretraining document is suitable for conversion into post-training data, such as pages consisting mostly of navigation links or product listings. As a tradeoff between quality and efficiency, we prompt Qwen3-4B~\citep{yang2025qwen3} to rate each candidate document on a ten-point quality scale and assign it to the factual knowledge, analytical reasoning, or creative writing category, or reject it as unsuitable (full prompt in Appendix Table~\ref{tab:source-selection-prompt}). We score randomly sampled documents until the top-rated count reaches a preset quota per category. Format following relies more on document structure than on semantics, so a fastText classifier~\citep{bojanowski2017fasttext}, trained with 300K instruction format following responses as positives and 600K random pretraining documents as negatives, first narrows the pool to the top one million documents, and the 50K rated highest by Qwen3-4B form the source pool.

\paragraph{QA construction with lexical constraints} To avoid introducing additional information, we apply a lexical constraint during QA pair construction. This builds on the finding from neural text summarisation~\citep{see2017pointer-generator} that models copy most words from the source document and add only simple connecting words, so coherent output can be composed almost entirely of source words. The constraint therefore allows each QA pair to contain only content from its source document and words from a predefined function word set.\footnote{The set covers particles, conjunctions, pronouns, and similar connectives that add no factual content.} Under this constraint, we prompt DeepSeek-V4-Pro~\citep{xu2026deepseekv4} to convert each selected document into QA pairs. As language models may deviate from the required format or hallucinate, we verify each pair with a rule-based script and keep only those that pass. The retained pairs thus contain no information beyond the source document and its cutoff date. We run this curation process independently for each of the twelve years from 2013 to 2024. The final dataset contains approximately 60K QA pairs per year, on par with Alpaca's 52K~\citep{alpaca}.

\section{Time-Aware Model Training}
\subsection{Training setups}
\paragraph{Model architecture and hyperparameters} We pretrain a series of LLMs from scratch using our curated pretraining set (detailed in Section~\ref{sec:pretrain-data}) with a strict date cutoff, referred to as \textsc{DatedGPT}. \textsc{DatedGPT} consists of 1.3B parameters in total, with an architecture and hyperparameters (detailed in Appendix Table~\ref{tab:model-arch}) following the Llama  and GPT-2 reproductions~\citep{karpathy2024llmc}.

\paragraph{Pretraining setup} We pretrain each model from scratch, producing 12 models with cutoff dates from 2013 to 2024. Following \citet{karpathy2024llmc}, we set the budget to 25,000 iterations, approximately 100B tokens per model, a reasonable tradeoff between performance and computational cost. Full architecture and training hyperparameters are listed in Appendix Table~\ref{tab:model-arch}.
Training curves are smooth throughout (Appendix Figure~\ref{fig:train_loss}), without the loss spikes reported in prior large-scale runs \citep{zhang2022opt, zeng2022glm, chowdhery2023palm}. 
Each round of pretraining requires approximately 2,000 GPU hours on A100 GPUs. 
To our knowledge, this ranks among the largest and most complete yearly coverage model series to date.\footnote{\citet{NBERw35247-pit} introduce a 4B-parameter point-in-time model series with a similar approach. This work was released less than two months before our submission and therefore qualifies as concurrent work under the ACL guidelines.}

\paragraph{Instruction-Tuning Setup}\label{sec:instruct-setup}
We introduce a second stage to enhance the models' instruction-following capability. Each yearly model is tuned for three epochs on \textsc{DatedInstruct}, our instruction dataset curated from that year's pretraining documents under strict temporal constraints (Section~\ref{sec:instruction-data}). To stabilise tuning and prevent catastrophic forgetting~\citep{bethune2025scaling-replay}, we mix 10\% of pretraining data into the tuning process.

\section{Experiment Results}
This section evaluates \textsc{DatedGPT} from three angles. We first validate general language understanding, comparing the series against open-source models of similar size and against other models trained with explicit time cutoffs, on standard benchmarks and pairwise win rates (AlpacaEval~2.0). We next validate the quality of \textsc{DatedInstruct} by tuning the base checkpoints of other time-cutoff model families on it, which yields consistent improvements over their original instruction-tuned versions. Finally, we turn to stock return prediction, establishing the predictive power of our models, quantifying the impact of lookahead bias in financial tasks, and visualising the perplexity reversal at each cutoff date.

\begin{table*}[t]
\centering
\scalebox{0.75}{
\begin{tabular}{lcccccccc}
\toprule
\textbf{Model (0-shot)} & \textbf{ARC-C} & \textbf{ARC-E} & \textbf{HellaSwag} & \textbf{PIQA} & \textbf{IFEval} & \textbf{MMLU} & \textbf{TruthfulQA} & \textbf{Avg.} \\
\midrule
\multicolumn{9}{l}{\textit{Baseline Models}} \\
\midrule
SmolLM-1.7B-Instruct & 34.9 & 54.3 & 56.1 & 72.4 & 15.5 & 27.1 & 27.7 & 41.1 \\
GPT2-XL & 28.4 & 51.0 & 50.8 & 70.5 & 15.0 & 25.3 & 22.4 & 37.6 \\
TinyLlama-1.1B & 30.5 & 43.7 & 55.0 & 72.5 & 14.8 & 24.8 & 26.0 & 38.2 \\
OPT-1.3B & 27.8 & 51.3 & 53.7 & 70.9 & 17.6 & 25.2 & 23.8 & 38.6 \\
Pythia-1B & 27.1 & 49.0 & 47.1 & 69.3 & 16.8 & 23.1 & 23.6 & 36.6 \\
\midrule
\multicolumn{9}{l}{\textit{Time-locked Models}} \\
\midrule
ChronoGPT-Instruct-2024 & 27.6 & 45.4 & 40.7 & 65.0 &  18.7 & 24.8 & 27.1 & 35.6 \\
ChronoGPT-2024 + \textsc{DatedInstruct} & 29.8 & 46.3 & 42.4 & 67.8 & 21.6 & 23.0 & 26.4 & 36.8 \\
PIT-Instruct-2024 & 30.6 & 46.3 & 63.3 & 74.6 & 19.2 & 23.1 & 25.3 & 40.3 \\
PIT-2024 + \textsc{DatedInstruct} & 34.5 & 52.7 & 66.6 & 78.3 & 30.1 & 23.0 & 23.0 & 44.0 \\
\midrule
\multicolumn{9}{l}{\textit{\textsc{DatedGPT-Instruct} Series}} \\
\midrule
\textsc{DatedGPT-Instruct-2013} & 32.4 & 45.9 & 47.4 & 68.9 & 34.9 & 23.1 & 23.5 & 39.4 \\
\textsc{DatedGPT-Instruct-2017} & 32.7 & 47.9 & 50.7 & 70.4 & 33.8 & 23.1 & 21.9 & 40.1 \\
\textsc{DatedGPT-Instruct-2024} & 34.6 & 49.6 & 53.6 & 70.7 & 35.3 & 23.7 & 24.9 & 41.8 \\
\bottomrule
\end{tabular}
}
\caption{Zero-shot evaluation results on language understanding benchmarks. IFEval reports prompt-level strict accuracy. The \textsc{DatedGPT-Instruct} series models are identified by their training data cutoff year. We report four representative models here, and the full series performance is detailed in Appendix~\ref{sec:appendix}. For each time-locked model, we use the latest available checkpoint of its family and report both the original instruction-tuned release and the base version tuned on \textsc{DatedInstruct}.}
\label{tab:main-results-instruct}
\end{table*}

\subsection{Evaluation setup}
\paragraph{Baseline and Time-locked models} We compare \textsc{DatedGPT} against several publicly available models of similar scale, including GPT-XL~\citep{radford2019language}, OPT-1.3B~\citep{zhang2022opt}, Pythia-1B~\citep{biderman2023pythia}, TinyLlama-1.1B~\citep{zhang2024tinyllama} and smolLM-1.7B~\citep{allal2025smollm2}. We also compare with models trained under explicit time cutoffs, including ChronoGPT~\citep{he2025chronologically} and PIT~\citep{NBERw35247-pit}, a 4B-parameter series with a similar design.

\paragraph{Our models} Our model development consists of two stages: general-purpose pretraining and time-aware instruction tuning. We denote the resulting models as \textsc{DatedGPT-base} and \textsc{DatedGPT-instruct}, respectively. Each model is identified by a year suffix to indicate the temporal cutoff of the training data (e.g., \textsc{DatedGPT-base-2024}).

\paragraph{Evaluation datasets} We evaluate a collection of widely used benchmarks for general language understanding performance, covering commonsense reasoning, scientific knowledge, factual accuracy, instruction following, and multitask academic knowledge. Specifically, we evaluate on HellaSwag~\citep{zellers2019hellaswag}, PIQA~\citep{bisk2020piqa}, ARC~\citep{clark2018arc}, TruthfulQA~\citep{lin2022truthfulqa}, IFEval~\citep{zhou2023ifeval}, and MMLU~\citep{hendrycks2020mmlu}. 

\subsection{Language Understanding Evaluation}

We report the zero-shot evaluation performance of \textsc{DatedGPT-Instruct} in this section. The base model results are provided in the Appendix.%

\paragraph{\textsc{DatedGPT} achieves competitive performance at small scale} We report the evaluation results in Table~\ref{tab:main-results-instruct}. Our \textsc{DatedGPT-Instruct} series models demonstrate consistently strong performance across a wide range of tasks, achieving an average score of up to 41.8. This places our models among the strongest small-scale instruction-following models available. Notably, the substantial gains on the IFEval benchmark indicate effective instruction-following behaviour. On general language understanding tasks such as ARC and HellaSwag, our models achieve competitive performance with leading small-scale models, despite being trained on significantly fewer resources.%

\paragraph{\textsc{DatedGPT} models achieve comparable performance across different years} Despite being trained with different cutoff years, our \textsc{DatedGPT-Instruct} models exhibit similar performance, with average scores ranging from 39.4 to 41.8. This suggests that models without access to up-to-date data can still develop strong general language understanding capabilities.

\paragraph{\textsc{DatedGPT} matches or surpasses time-locked models on pairwise win rates} Beyond benchmark accuracy, we compare the models through pairwise evaluation following AlpacaEval~2.0. Table~\ref{tab:win-rate} reports the win rates of \textsc{DatedGPT-Instruct-2024} against the 2024 version of each time-locked family. \textsc{DatedGPT} wins significantly more head-to-head comparisons than ChronoGPT-Instruct, with 81.1\% wins against 18.3\% losses, and performs on par with PIT-Instruct, winning 46.6\% of the comparisons against 52.9\% losses, despite being three times smaller.

\begin{table}[t]
\centering
\small
\begin{tabular}{lccc}
\toprule
\textbf{Opponent} & \textbf{Win} & \textbf{Tie} & \textbf{Loss} \\
\midrule
ChronoGPT-Instruct & 81.1 & 0.62 & 18.3 \\
PIT-Instruct & 46.6 & 0.50 & 52.9 \\
\bottomrule
\end{tabular}
\caption{Win, tie, and loss rates (\%) of \textsc{DatedGPT-Instruct} against each time-locked family's 2024 instruct version, following AlpacaEval~2.0.}
\label{tab:win-rate}
\end{table}

\begin{table*}[!ht]
\centering
\scalebox{0.76}{
\begin{tabular}{lrrrcrrrcc}
\toprule
& \multicolumn{4}{c}{\textbf{Panel regression}} & \multicolumn{5}{c}{\textbf{Portfolio}} \\
\cmidrule(lr){2-5} \cmidrule(lr){6-10}
& & & & \textbf{LH-B} & & & & & \textbf{LH-B} \\
\textbf{Model} & $\hat\beta$ \textbf{(pp)} & $t$ & $\hat\beta\!\cdot\!\textbf{SD (bp)}$ & $>$\textbf{LH-F} & \textbf{Sharpe} & $\mu_\mathrm{LS}$ & $\alpha_\mathrm{FF5}$ & $t(\alpha)$ & $>$\textbf{LH-F} \\
\midrule
\multicolumn{10}{l}{\textit{Panel A. Lookahead-free (vintage $y{-}1$ scores year $y$)}} \\
\midrule
\multicolumn{10}{l}{\quad\textit{Base models:}} \\
PIT-4B             & $0.331$ & $13.30$ & $29.6$ & $\checkmark$ & $3.37$ & $0.57$ & $0.56$ & $10.16$ & $\checkmark$ \\
\textsc{DatedGPT}-base      & $0.112$ & $2.64$  & $4.9$  & $\checkmark$ & $0.78$ & $0.09$ & $0.09$ & $2.48$  & $\checkmark$ \\
ChronoGPT-base     & $0.007$ & $0.06$  & $0.0$  & $\times$     & $0.18$ & $0.04$ & $0.00$ & $0.54$  & $\times$     \\
\multicolumn{10}{l}{\quad\textit{Instruct models:}} \\
\textsc{DatedGPT}-instruct  & $0.443$ & $11.65$ & $28.3$ & $\checkmark$ & $3.20$ & $0.73$ & $0.72$ & $10.09$ & $\checkmark$ \\
PIT-4B-FT          & $0.171$ & $2.62$  & $5.9$  & $\checkmark$ & $0.86$ & $0.12$ & $0.12$ & $2.74$  & $\checkmark$ \\
ChronoGPT-Instruct & $0.071$ & $2.19$  & $5.8$  & $\times$     & $0.59$ & $0.13$ & $0.13$ & $2.09$  & $\checkmark$ \\
\midrule
\multicolumn{10}{l}{\textit{Panel B. Off-the-shelf lookahead-biased baselines}} \\
\midrule
\multicolumn{10}{l}{\quad\textit{Base models:}} \\
Llama-2-7B         & $0.577$ & $15.95$ & $35.3$ & --- & $4.01$ & $1.02$ & $1.02$ & $12.62$ & --- \\
Gemma-3-4B         & $0.271$ & $13.79$ & $26.9$ & --- & $3.91$ & $0.49$ & $0.49$ & $11.74$ & --- \\
Llama-3.2-1B       & $0.289$ & $7.64$  & $24.6$ & --- & $3.30$ & $0.48$ & $0.48$ & $10.86$ & --- \\
Llama-3.2-3B       & $0.511$ & $7.11$  & $21.5$ & --- & $2.77$ & $0.55$ & $0.55$ & $8.80$  & --- \\
\multicolumn{10}{l}{\quad\textit{Instruct models:}} \\
Llama-3.2-3B-Instruct & $0.357$ & $9.02$  & $29.5$ & --- & $3.75$ & $0.61$ & $0.61$ & $12.53$ & --- \\
Gemma-3-4B-it      & $0.279$ & $13.66$ & $26.1$ & --- & $3.82$ & $0.49$ & $0.50$ & $12.06$ & --- \\
Llama-3.2-1B-Instruct & $0.251$ & $8.20$  & $24.9$ & --- & $3.60$ & $0.45$ & $0.44$ & $11.21$ & --- \\
Llama-2-7B-chat    & $0.234$ & $7.91$  & $13.4$ & --- & $1.63$ & $0.30$ & $0.30$ & $5.33$  & --- \\
\midrule
\multicolumn{10}{l}{\textit{Panel C. $v_{2024}$ lookahead (latest vintage scores all years)}} \\
\midrule
\multicolumn{10}{l}{\quad\textit{Base models:}} \\
PIT-4B ($v_{2024}$)            & $0.584$ & $13.91$ & $32.7$ & --- & $3.52$ & $0.90$ & $0.90$ & $11.52$ & --- \\
\textsc{DatedGPT}-base ($v_{2024}$)      & $0.147$ & $3.38$  & $5.9$  & --- & $1.27$ & $0.24$ & $0.24$ & $3.94$  & --- \\
ChronoGPT-base ($v_{2024}$)   & $0.032$ & $0.85$  & $0.0$  & --- & $0.18$ & $0.05$ & $0.00$ & $0.43$  & --- \\
\multicolumn{10}{l}{\quad\textit{Instruct models:}} \\
\textsc{DatedGPT}-instruct ($v_{2024}$)  & $0.617$ & $14.86$ & $33.0$ & --- & $3.65$ & $1.02$ & $1.03$ & $10.37$ & --- \\
PIT-4B-FT ($v_{2024}$)        & $0.240$ & $4.05$  & $9.7$  & --- & $1.46$ & $0.40$ & $0.39$ & $4.60$  & --- \\
ChronoGPT-Instruct ($v_{2024}$) & $0.189$ & $1.68$  & $0.0$  & --- & $1.22$ & $0.32$ & $0.32$ & $2.52$  & --- \\
\bottomrule
\end{tabular}
}
\caption{Panel regression and portfolio results for stock return prediction using LLM-based sentiment signals. See Section~\ref{sec:stock-prediction} for variable definitions and setup.}
\label{tab:stock-prediction}
\end{table*}

\paragraph{\textsc{DatedInstruct} leads to consistent performance improvements for time-locked models} To assess whether \textsc{DatedInstruct} is useful beyond our own models, we apply it to two other model families with explicit date cutoffs, ChronoGPT~\citep{he2025chronologically} and PIT~\citep{NBERw35247-pit}. We take each family's base checkpoint, tune it on \textsc{DatedInstruct} with the setup described in Section~\ref{sec:instruct-setup}, and compare the result against the family's own instruction-tuned release. The time-locked group of Table~\ref{tab:main-results-instruct} reports the zero-shot results. Tuning with \textsc{DatedInstruct} consistently improves over each family's original instruction-tuned version, raising the PIT average by 10\%.

\subsection{Stock Return Prediction and Lookahead Bias}\label{sec:stock-prediction}

We use 61,290 firm-day stock news headlines from 2014 to 2024. Each model is prompted with: \texttt{``Classify this news headline as either good, or bad, or neutral for the stock price of company. Headline: \{headline\}''} and the response is mapped to $s_{i,t} \in \{+1, 0, -1\}$ for good, neutral, and bad. Following \citet{lopez2023can}, we estimate $r_{i,t+1} = \alpha_i + \delta_t + \beta\, s_{i,t} + \varepsilon_{i,t+1}$ with firm and date fixed effects and standard errors clustered by date, where $r_{i,t+1}$ is the next-day stock return. $\hat\beta\!\cdot\!\mathrm{SD}$ is the economic magnitude in basis points, the change in return per one-standard-deviation increase in the signal. We also form a daily equal-weighted long-short portfolio (long good, short bad). $\alpha_{\mathrm{FF5}}$ and $t(\alpha)$ are the daily alpha and $t$-statistic with respect to \citet{fama2015five}. The LH-B~$>$~LH-F columns indicate whether the $v_{2024}$ lookahead result (Panel~C) exceeds the lookahead-free result (Panel~A), comparing $\hat\beta\!\cdot\!\mathrm{SD}$ for the regression and the Sharpe ratio for the portfolio.

\paragraph{Lookahead-bias-free models predict stock returns} Panel~A of Table~\ref{tab:stock-prediction} shows that lookahead-bias-free models earn positive and significant returns. Among base models, PIT-4B achieves $29.6$~bp per SD, slightly below the lookahead-biased Llama-2-7B ($35.3$~bp) but exceeding Gemma-3-4B ($26.9$~bp). Among instruct models, \textsc{DatedGPT}-instruct achieves $28.3$~bp, comparable to Llama-3.2-3B-Instruct ($29.5$~bp) and larger than Gemma-3-4B-it ($26.1$~bp), despite being 1.3B parameters with a strict temporal cutoff. These results show that lookahead-bias-free models are useful for applications such as news-based prediction.

\paragraph{Lookahead bias inflates predictive performance} Comparing Panels~A and~C, the lookahead-biased models consistently outperform their lookahead-free counterparts: PIT-4B improves from $29.6$ to $32.7$~bp and from $3.37$ to $3.52$ in Sharpe; \textsc{DatedGPT}-instruct from $28.3$ to $33.0$~bp and $3.20$ to $3.65$; PIT-4B-FT from $5.9$ to $9.7$~bp and $0.86$ to $1.46$. ChronoGPT does not show this pattern, consistent with its weak overall performance. This improvement is unlikely driven by data quality: PIT's lookahead-free vintages show stable performance after 2014, and \textsc{DatedGPT} retrains from scratch on ${\sim}$100B tokens per vintage.

To formally test this, Table~\ref{tab:lookahead-premium} decomposes the lookahead-biased signal into a lookahead-free component and a residual. Let $s^{f}_{i,t}$ denote the lookahead-free signal from vintage $y{-}1$ scoring year $y$ headlines, and $s^{b}_{i,t}$ the lookahead-biased signal from the family's latest vintage applied to all headlines. We write $s^{b} = s^{f} + (s^{b} - s^{f})$, where the residual $(s^{b} - s^{f})$ isolates the additional information available only to the lookahead-biased vintage. We then estimate $r_{i,t+1} = \alpha_i + \delta_t + \beta_1 s^{f} + \beta_2 (s^{b} - s^{f}) + \varepsilon$, where a positive and significant $\hat\beta_2$ indicates that the lookahead-biased vintage adds predictive content beyond the lookahead-free signal. Among base models, PIT-4B shows a large premium of $40.1$~bp ($t = 10.98$) and \textsc{DatedGPT}-base $7.7$~bp ($t = 3.13$). Among instruct models, \textsc{DatedGPT}-instruct yields $26.4$~bp ($t = 10.65$) and PIT-4B-FT $12.2$~bp ($t = 3.99$). ChronoGPT yields insignificant $\hat\beta_2$, consistent with its weak baseline performance.

\begin{table}[!ht]
\centering
\small
\scalebox{0.78}{
\begin{tabular}{lrrr}
\toprule
\textbf{Family} & $\hat\beta_2$ \textbf{(pp)} & $t(\hat\beta_2)$ & $\hat\beta_2\!\cdot\!\textbf{SD (bp)}$ \\
\midrule
\multicolumn{4}{l}{\textit{Panel A. Base models}} \\
\midrule
PIT-4B             & $0.465$ & $10.98$ & $40.1$ \\
\textsc{DatedGPT}-base      & $0.138$ & $3.13$  & $7.7$  \\
ChronoGPT-base     & $0.033$ & $0.90$  & $0.0$  \\
\midrule
\multicolumn{4}{l}{\textit{Panel B. Instruct models}} \\
\midrule
\textsc{DatedGPT}-instruct  & $0.472$ & $10.65$ & $26.4$ \\
PIT-4B-FT          & $0.236$ & $3.99$  & $12.2$ \\
ChronoGPT-Instruct & $0.172$ & $1.51$  & $0.0$  \\
\bottomrule
\end{tabular}
}
\caption{Lookahead premium test. See Section~\ref{sec:stock-prediction} for the full specification.}
\label{tab:lookahead-premium}
\end{table}

\begin{figure}[h]
\centering
\includegraphics[width=0.8\columnwidth]{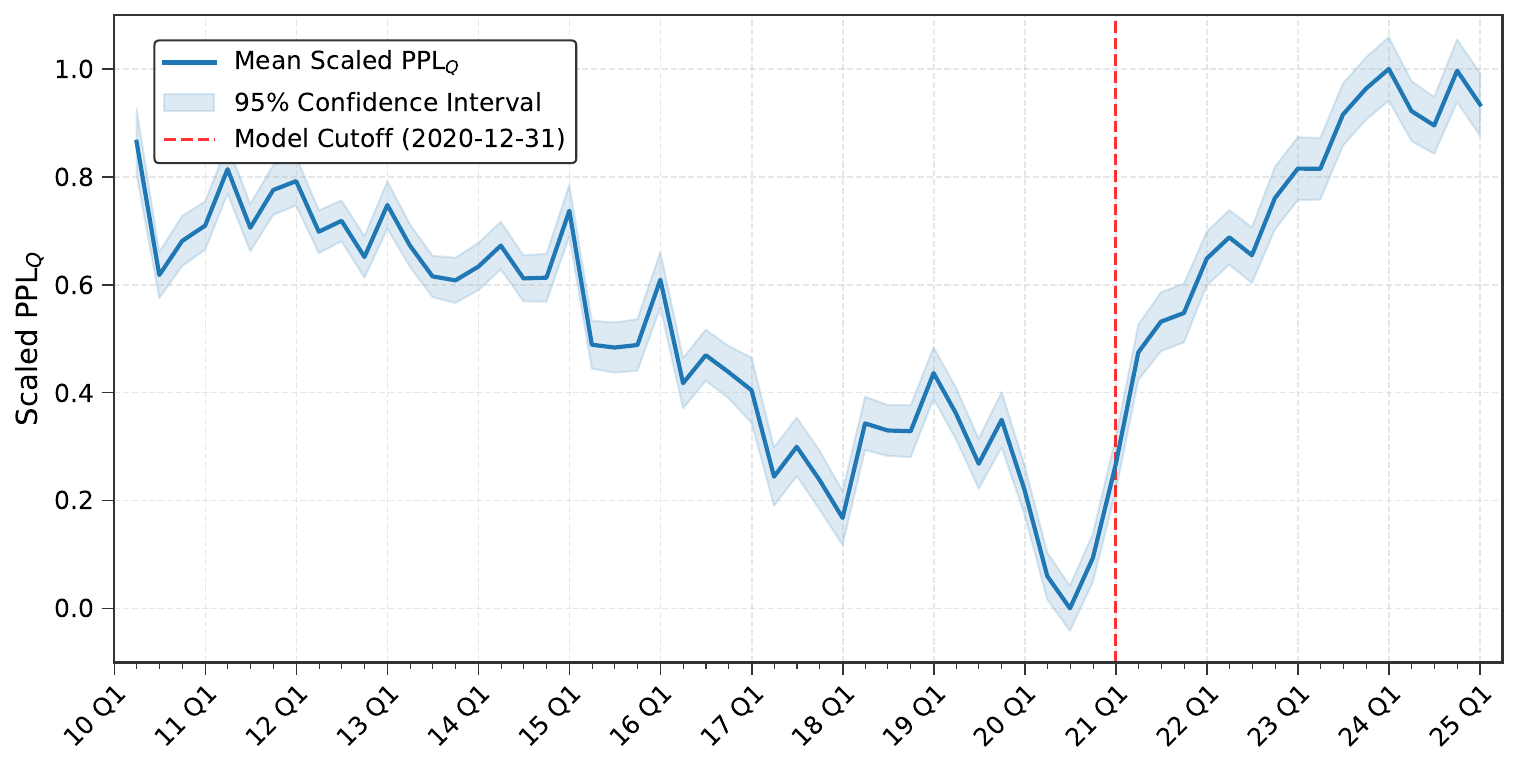}
\caption{Average relative perplexity of \textsc{DatedGPT-base-2020} evaluated on quarterly public company news headlines from 2013 to 2024.}
\label{fig:cutoff-vis}
\end{figure}

\paragraph{Knowledge memorisation visualisation}\label{section:memorisation} To verify that each model's knowledge is bounded by its cutoff date, we adopt the perplexity-based probing method of \citet{cheng2024dated} and evaluate the \textsc{DatedGPT-base} series on quarterly news headlines of publicly listed companies~\citep{gao2025test}. Figure~\ref{fig:cutoff-vis} shows the results for \textsc{DatedGPT-base-2020}. A clear reversal appears around the model's 2020 data cutoff, after which perplexity rises steadily, confirming that the model has not been exposed to post-cutoff text. The same pattern holds across all cutoff years.

\bibliography{custom}

\clearpage

\appendix
\section{Dataset Details}
\begin{table}[t]
\centering
\begin{tabular}{p{0.94\columnwidth}}
\toprule
\textbf{Source-Document Evaluation Prompt} \\
\midrule
{\footnotesize\ttfamily\raggedright
\textbf{System:} You are a data quality evaluator for SFT (Supervised Fine-Tuning) dataset creation. Given a document, determine whether it is suitable to be rewritten as SFT data and output a JSON object with exactly two fields:\par
-- "score": integer 1-10. 1=garbage/SEO/truncated/chaotic, 10=perfect.\par
-- "category": one of ["reasoning / logical-analysis", "knowledge / factual-QA", "writing / creative", "low-quality / unsuitable"].\par
Consider: information density, clarity, coherence, factual content, educational value, and whether the text could support good training examples. Output ONLY the JSON object. No other text.\par
\vspace{4pt}
\textbf{User:} Document: \{text\}\par
Evaluate. Output JSON only.\par
} \\
\bottomrule
\end{tabular}
\caption{Prompt used for source-document selection. Qwen3-4B rates each candidate document on a 1--10 quality scale and assigns it to one of the three content categories or rejects it as unsuitable.}
\label{tab:source-selection-prompt}
\end{table}

\section{Model Training Details}
\begin{table}[h]
\begin{center}
\small 
\centering
\setlength{\tabcolsep}{8pt}
\resizebox{0.9\columnwidth}{!}{%
\begin{tabular}{lc}
\toprule
 Hyperparameter &  Value\\
\midrule
Sequence Length &   2048 \\
    Number of Layers &    24 \\
   Embedding Size &    2048 \\
   FFN Hidden Size &   5504 \\
   Number of Heads &   16 \\
   Position Encodings &   RoPE \\
   Activation Function &   SwiGLU\\
   Layer Norm &   RMSNorm \\
   Learning Rate &   2E-4 \\
   Batch Size &   2048 \\
   Vocabulary Size &    32000 \\
 \midrule
   Embedding Parameters &   0.13B \\
   Non-Embedding Parameters &   1.21B\\
   Total Parameters &   1.34B\\
\bottomrule
\end{tabular}
}
\end{center}
\caption{Model and pretraining hyperparameters.} 
\label{tab:model-arch}
\end{table}

\begin{figure}[h]
\centering
\includegraphics[width=\columnwidth]{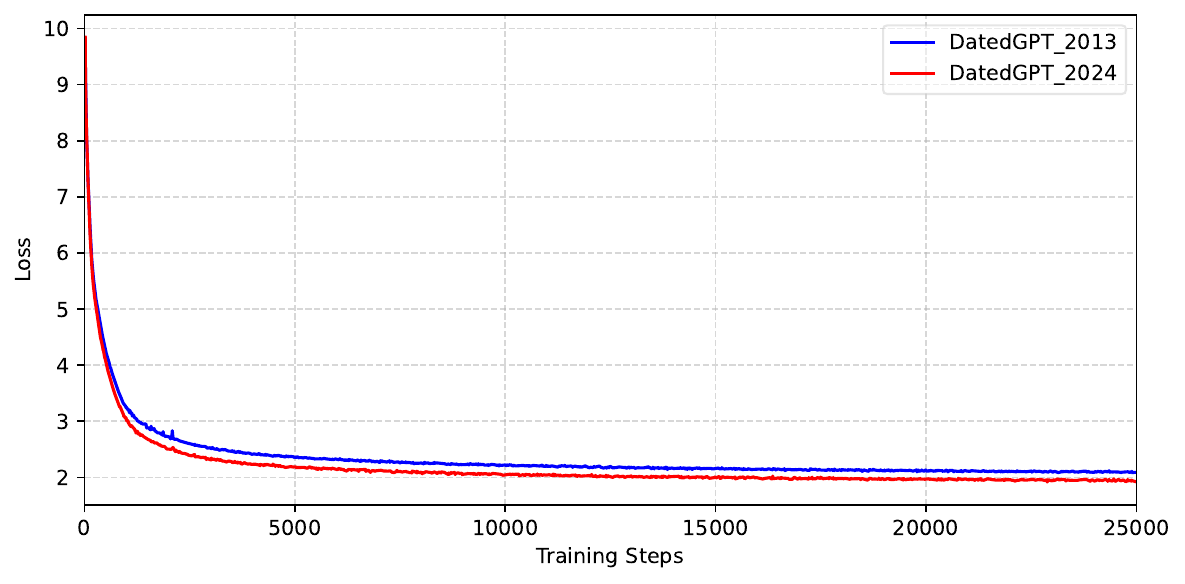}
\caption{Training loss curves for 2013 and 2024 \textsc{DatedGPT-base} models. All models exhibit smooth convergence without loss spikes or instability.}
\label{fig:train_loss}
\end{figure}%

\begin{figure}[h]
\centering
\includegraphics[width=\columnwidth]{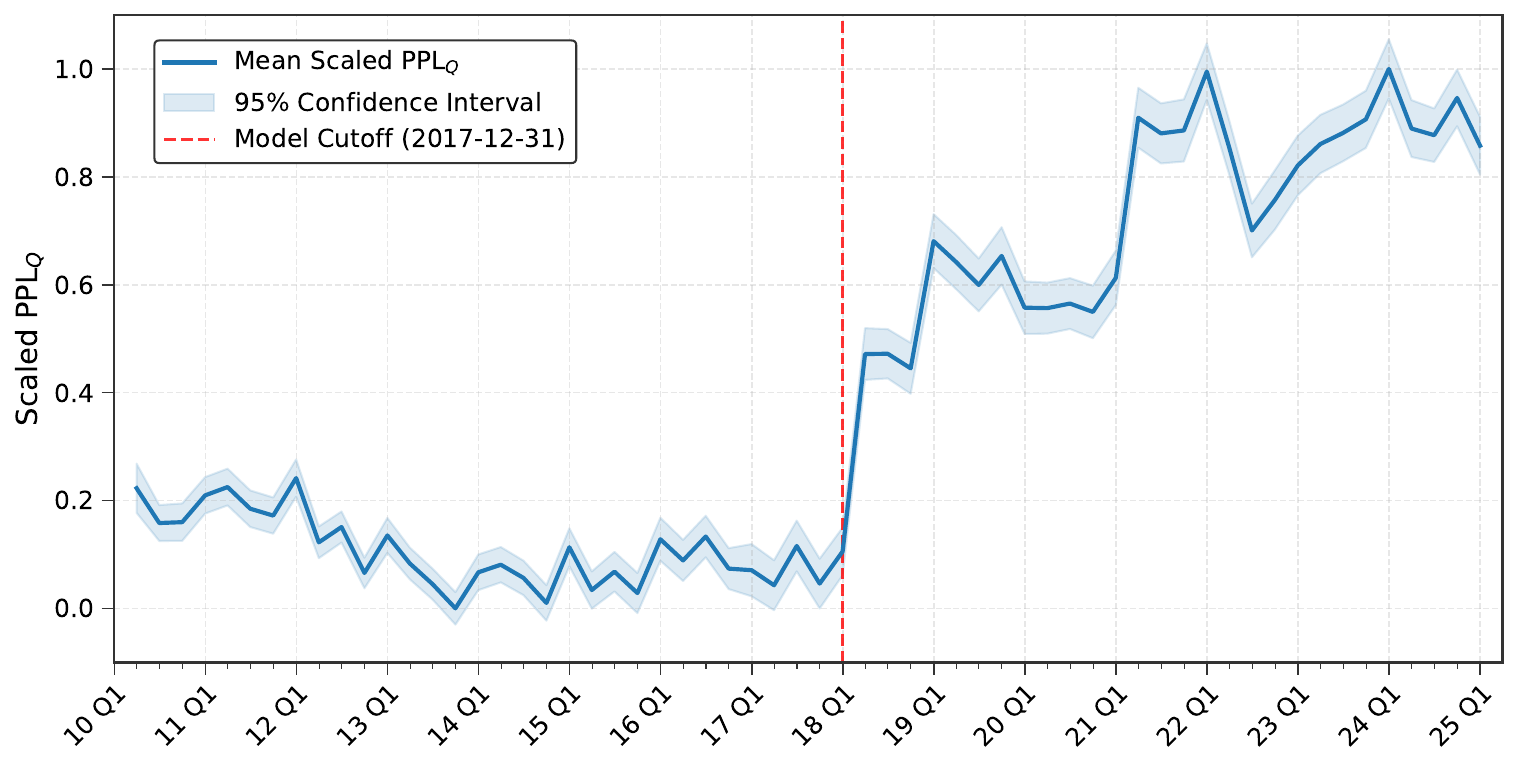}
\caption{Relative perplexity of \textsc{DatedGPT-base-2017} evaluated on quarterly public company news head-lines from 2013 to 2024.}
\label{fig:cutoff-vis-2017}
\end{figure}%

\section{Evaluation Result on Full Series of \textsc{DatedGPT}}
\label{sec:appendix}
\begin{table*}[t]
\centering
\scalebox{0.9}{
\begin{tabular}{lccccccc}
\toprule
\textbf{Model (0-shot)} & \textbf{ARC-C} & \textbf{ARC-E} & \textbf{HellaSwag} & \textbf{PIQA} & \textbf{IFEval} & \textbf{MMLU} & \textbf{TruthfulQA} \\
\midrule
\multicolumn{8}{l}{\textit{Baseline Models}} \\
\midrule
SmolLM-1.7B-Instruct & 34.9 & 54.3 & 56.1 & 72.4 & 15.5 & 27.1 & 27.7 \\
GPT2-XL & 28.4 & 51.0 & 50.8 & 70.5 & 15.0 & 25.3 & 22.4 \\
TinyLlama-1.1B & 30.5 & 43.7 & 55.0 & 72.5 & 14.8 & 24.8 & 26.0 \\
OPT-1.3B & 27.8 & 51.3 & 53.7 & 70.9 & 17.6 & 25.2 & 23.8 \\
Pythia-1B & 27.1 & 49.0 & 47.1 & 69.3 & 16.8 & 23.1 & 23.6 \\
\midrule
\multicolumn{8}{l}{\textit{\textsc{DatedGPT-Instruct} Series}} \\
\midrule
\textsc{DatedGPT-Instruct-2013} & 33.6 & 48.8 & 47.6 & 66.4 & 34.2 & 25.3 & 25.1 \\
\textsc{DatedGPT-Instruct-2014} & 32.6 & 47.3 & 47.8 & 69.2 & 34.4 & 24.8 & 24.7 \\
\textsc{DatedGPT-Instruct-2015} & 32.3 & 48.0 & 48.0 & 67.1 & 34.8 & 24.1 & 26.6 \\
\textsc{DatedGPT-Instruct-2016} & 32.9 & 50.6 & 48.4 & 69.4 & 32.5 & 24.8 & 26.6 \\
\textsc{DatedGPT-Instruct-2017} & 34.8 & 50.5 & 50.0 & 69.0 & 34.0 & 24.8 & 26.8 \\
\textsc{DatedGPT-Instruct-2018} & 35.2 & 50.0 & 50.5 & 68.9 & 30.7 & 25.3 & 27.2 \\
\textsc{DatedGPT-Instruct-2019} & 31.1 & 49.8 & 50.6 & 69.5 & 34.0 & 24.9 & 25.3 \\
\textsc{DatedGPT-Instruct-2020} & 33.1 & 48.5 & 51.1 & 68.9 & 32.5 & 25.7 & 25.1 \\
\textsc{DatedGPT-Instruct-2021} & 32.9 & 48.5 & 51.0 & 68.9 & 31.8 & 24.4 & 26.4 \\
\textsc{DatedGPT-Instruct-2022} & 33.6 & 49.9 & 51.4 & 69.2 & 35.9 & 23.5 & 26.9 \\
\textsc{DatedGPT-Instruct-2023} & 32.3 & 51.1 & 52.7 & 69.7 & 33.8 & 23.7 & 30.8 \\
\textsc{DatedGPT-Instruct-2024} & 34.7 & 52.0 & 53.2 & 70.5 & 35.3 & 24.3 & 28.6 \\
\bottomrule
\end{tabular}
}
\caption{Zero-shot evaluation results on language understanding benchmarks. IFEval reports prompt-level strict accuracy. The \textsc{DatedGPT-Instruct} series models are identified by their training data cutoff year.}
\label{tab:main-results-instruct-appendix}
\end{table*}

\begin{table*}[t]
\centering
\scalebox{0.9}{
\begin{tabular}{lccccccc}
\toprule
\textbf{Model (5-shot)} & \textbf{ARC-C} & \textbf{ARC-E} & \textbf{HellaSwag} & \textbf{PIQA} & \textbf{IFEval} & \textbf{MMLU} & \textbf{TruthfulQA} \\
\midrule
\multicolumn{8}{l}{\textit{Baseline Models}} \\
\midrule
SmolLM-1.7B-Instruct & 43.9 & 74.0 & 62.4 & 74.1 & 4.6 & 28.3 & 25.3 \\
GPT2-XL & 29.9 & 59.4 & 51.2 & 70.6 & 15.0 & 26.3 & 22.0 \\
TinyLlama-1.1B & 35.5 & 66.4 & 60.8 & 74.8 & 3.1 & 25.0 & 23.6 \\
OPT-1.3B & 29.2 & 59.6 & 54.1 & 71.0 & 17.4 & 24.8 & 24.0 \\
Pythia-1B & 28.2 & 56.9 & 47.5 & 70.1 & 17.4 & 25.8 & 23.6 \\
\midrule
\multicolumn{8}{l}{\textit{\textsc{DatedGPT-Base} Series}} \\
\midrule
\textsc{DatedGPT-Base-2013} & 34.6 & 67.2 & 48.8 & 69.8 & 17.6 & 25.1 & 21.9 \\
\textsc{DatedGPT-Base-2014} & 37.0 & 67.0 & 48.8 & 69.7 & 17.0 & 25.5 & 22.0 \\
\textsc{DatedGPT-Base-2015} & 36.0 & 68.2 & 49.0 & 70.2 & 14.4 & 24.7 & 21.1 \\
\textsc{DatedGPT-Base-2016} & 37.8 & 67.9 & 49.3 & 69.2 & 12.6 & 25.5 & 20.7 \\
\textsc{DatedGPT-Base-2017} & 39.3 & 69.3 & 51.3 & 71.1 & 19.8 & 25.3 & 21.9 \\
\textsc{DatedGPT-Base-2018} & 38.3 & 68.7 & 51.8 & 71.0 & 15.7 & 25.3 & 20.9 \\
\textsc{DatedGPT-Base-2019} & 35.0 & 67.7 & 51.6 & 71.7 & 14.6 & 24.2 & 20.0 \\
\textsc{DatedGPT-Base-2020} & 37.8 & 68.5 & 52.6 & 71.0 & 10.9 & 26.3 & 20.6 \\
\textsc{DatedGPT-Base-2021} & 38.4 & 68.6 & 52.3 & 70.5 & 12.8 & 25.8 & 22.9 \\
\textsc{DatedGPT-Base-2022} & 36.8 & 68.7 & 52.7 & 71.7 & 15.7 & 25.2 & 21.4 \\
\textsc{DatedGPT-Base-2023} & 39.8 & 70.0 & 53.6 & 71.8 & 13.9 & 23.7 & 23.9 \\
\textsc{DatedGPT-Base-2024} & 40.2 & 71.6 & 54.6 & 71.3 & 12.8 & 25.8 & 24.4 \\
\bottomrule
\end{tabular}
}
\caption{Five-shot evaluation results on language understanding benchmarks. IFEval reports prompt-level strict accuracy. The \textsc{DatedGPT-Base} series models are identified by their training data cutoff year.}
\label{tab:main-results-base-appendix}
\end{table*}

\end{document}